\documentclass[lettersize,journal]{IEEEtran}
\usepackage{amsmath,amsfonts}
\usepackage{algorithm}
\usepackage{algorithmic}
\usepackage{array}
\usepackage[caption=false,font=normalsize,labelfont=sf,textfont=sf]{subfig}
\usepackage{textcomp}
\usepackage{stfloats}
\usepackage{url}
\usepackage{verbatim}
\usepackage{graphicx}
\usepackage{xcolor}
\usepackage{cite}
\usepackage[normalem]{ulem}
\def\BibTeX{{\rm B\kern-.05em{\sc i\kern-.025em b}\kern-.08em
    T\kern-.1667em\lower.7ex\hbox{E}\kern-.125emX}}
\usepackage{balance}
\begin{document}
\title{SVInvNet: A Densely Connected Encoder-Decoder Architecture for Seismic Velocity Inversion}
\author{Mojtaba Najafi Khatounabad*, Hacer Yalim Keles, Selma Kadioglu

\thanks{%Manuscript created October, 2020; This work was developed by the IEEE Publication Technology Department. This work is distributed under the \LaTeX \ Project Public License (LPPL) ( http://www.latex-project.org/ ) version 1.3. A copy of the LPPL, version 1.3, is included in the base \LaTeX \ documentation of all distributions of \LaTeX \ released 2003/12/01 or later. The opinions expressed here are entirely that of the author. No warranty is expressed or implied. User assumes all risk.}

*Corresponding author: Mojtaba Najafi Khatounabad is with Institute of Science and Technology, Ankara University, Ankara, Türkiye. (email:najafim@ankara.edu.tr)}
\thanks{Hacer Yalim Keles is with the Department of Computer Engineering, Hacettepe University, Ankara, Türkiye (email: hacerkeles@cs.hacettepe.edu.tr}
\thanks{Selma Kadioglu is with the Department of Geophysical Engineering, Ankara University, Ankara, Türkiye (email:kadioglu@ankara.edu.tr)}}

\markboth{Journal of \LaTeX\ Class Files,~Vol.~18, No.~9, September~2020}%
{How to Use the IEEEtran \LaTeX \ Templates}

\maketitle

\begin{abstract}
This study presents a deep learning-based approach to seismic velocity inversion problem, focusing on both noisy and noiseless training datasets of varying sizes. Our Seismic Velocity Inversion Network (SVInvNet) introduces a novel architecture that contains a multi-connection encoder-decoder structure enhanced with dense blocks. This design is specifically tuned to effectively process time series data, which is essential for addressing the challenges of non-linear seismic velocity inversion. For training and testing, we created diverse seismic velocity models, including multi-layered, faulty, and salt dome categories. We also investigated how different kinds of ambient noise, both coherent and stochastic, and the size of the training dataset affect learning outcomes. SVInvNet is trained on datasets ranging from 750 to 6,000 samples and is tested using a large benchmark dataset of 12,000 samples. Despite its fewer parameters compared to the baseline model, SVInvNet achieves superior performance with this dataset.  The performance of SVInvNet was further evaluated using the OpenFWI dataset and Marmousi-derived velocity models. The comparative analysis clearly reveals the effectiveness of the proposed model. 

\end{abstract}

\begin{IEEEkeywords}
Seismic velocity inversion, deep learning, encoder decoder architecture, convolutional neural network, CNN, densenet.
\end{IEEEkeywords}

\section{Introduction}
\IEEEPARstart{T}{he} seismic reflection method is widely utilized for mapping subsurface structures, and it plays a crucial role in the exploration of hydrocarbon reservoirs and geothermal energy sources. In the surface-to-surface seismic data acquisition technique, an artificial source on the surface generates seismic energy that is transmitted underground. As this seismic wavefield propagates beneath the surface, it undergoes reflection, refraction, and diffraction at interfaces with varying seismic velocity values, causing it to return to the surface. The returning waves are captured by receivers placed on the surface and recorded as a time series. These recordings, known as seismic traces, are aligned next to one another and create seismic shot gather. This assembly of seismic shot gathers is instrumental in probing and understanding the properties beneath the Earth's surface.

To extract depth-related information from time series seismic data, it's essential to have a subsurface velocity distribution that can accurately predict the observed seismic data. The effectiveness of various seismic imaging techniques, including migration as detailed by Baysal et al. \cite{Baysal} and Berkhout\cite{berkhout} and also inversion methods as noted by Scales \cite{Scales} and Schuster\cite{schuster2017seismic} depends on the precision of the velocity model. Although seismic data are crucial for developing the velocity model, their relationship is governed by the hyperbolic wave propagation equation, which is nonlinear. This complexity is compounded by the fact that different velocity models can yield identical seismic data, leading to an ill-posed inversion problem. This situation highlights the challenge of identifying a unique velocity model that aligns with the observed seismic data.

One common approach to addressing the challenges of subsurface velocity modeling and seismic data interpretation is through derivative-based inversion methods. These methods involve iteratively adjusting the parameters of an initial model to develop a velocity model that predict seismic data with a heightened resemblance to the observed data. This process includes overlaying synthetic and observed seismic data to determine the direction and magnitude of each adjustment. Among these methods, full waveform inversion (FWI) stands out as particularly effective, as noted in studies by Tarantola \cite{Taran84}, Gautheir et al.\cite{Gauthier}, Pan et al. \cite{Pan}, and Virieux and Operto\cite{Virieux}. FWI is distinguished by its use of both kinematic (arrival time) and dynamic (phase and amplitude) information from seismic data. However, it requires an initial velocity model that closely predicts the observed data, with differences less than half of the seismic wavelengths. If this condition isn't met, the method risks converging on local minimums due to cyclic-skipping problems, a challenge highlighted in research by Bunk et al.\cite{Bunk} and Sirgue and Pratt\cite{Sirgue}. 

In conjunction with ongoing investigations to address challenges associated with derivative-based methods, the effectiveness of innovative approaches is extensively examined. Significantly, within the realm of artificial intelligence advancements, deep neural networks stand out as a powerful tool for the mapping of seismic data to velocity model. The deep cascaded structure of these networks help learning complex non-linear functions from data. Through an iterative training process, the deep model adjusts the connections to capture the complex relationship between the seismic data and the velocity model. This approach allows the trained deep model to predict the velocity model accurately from seismic data, even if that data wasn't part of the training set \cite{Gabriel-Sarkar}, \cite{Bustamante-Fabien}. This advancement has the potential to reduce the reliance on initial seismic velocity models.

Due to the size and dimensional differences between seismic data and velocity model, the training process with deep models involves converting data from one domain to another. The encoder-decoder framework, which is good at deriving features from the input domain and predicting target in another domain, has demonstrated promising results in seismic inversion. In one of the earliest works, Wu et al.\cite{Wu} used this architecture and designed an end-to-end convolutional neural network (CNN) called InversionNet. InversionNet was able to predict velocity models with four-layer horizontal interface and fault with certain success. Within the encoder segment of the InversionNet, the size of the features undergo reduction to 1$\times$1, at the same time increasing the number of channels to 1024 progressively. 

Later, Li et al.\cite{Li} proposed SeisInvNet model using seismic data to construct velocity model. In SeisInvNet, which generally employs the encoder-decoder principle, each seismic trace is transformed into the features equivalent in size to the velocity model before entering the decoder section of the deep model. The improved version of SeisInvNet \cite{Liu} were successful to predict multilayer velocity models including geologic structures such as fault and salt dome with inclined and undulating interfaces. In this study, receiver gathers are employed in addition to shot gathers as input data. In these architectures, each layer is limited to utilize features exclusively from its immediate preceding layer.

To effectively train deep networks for learning the complex nonlinear functions we've discussed, it's crucial to have sufficient depth in the network. This depth is typically achieved by stacking multiple layers. In the training phase, the deep model attempts to estimate the velocity model by adjusting the weights in the layers. To facilitate this adjustment, it compares the result with the ground truth in each iteration and relays the difference through the layers to the first one using the backpropagation algorithm. Consequently, increasing the depth introduces a challenge; vanishing gradient problem \cite{Glorot10}. To tackle this problem, which is often encountered in deep CNN structures, we implement a variant of the DenseNet architecture, as introduced by Huang et al.\cite{Huang}. 

Unlike traditional CNN architectures, DenseNet uniquely features multiple connections between each layer and its subsequent layers in a series of dense blocks, which effectively prevents the fading of derivatives and accelerates information flow between the layers and dense blocks. The feature size in each dense block remains constant, with the inputs to each layer formed by concatenating the outputs of all preceding layers. We propose a novel encoder-decoder type architecture where both the encoder and the decoder parts are designed using dense blocks.

Our improved deep model, which we refer to as the Seismic Velocity Inversion Network (SVInvNet) furthermore, inspired from the UNet architecture, outputs from specific dense blocks in the encoder segment are connected to corresponding blocks in the decoder to support  the information flow. Thanks to the multiple connections among SVInvNet layers, the distance between input and ground truth is much shorter than in earlier designs.  We utilize raw seismic traces within a simple and straightforward architecture, without applying complex transformations to the input.

In order to use in our research, we manually prepared a novel dataset with 18,000 labeled data, pairing seismic shot gathers with velocity models. To mimic real data more closely, we introduced both coherent and stochastic noise to the synthetic data, resulting in an additional 18,000 noisy labeled data. We experimented with five training datasets, with varying sizes from 750 to 6,000. We assessed the impact of the data volume on our deep model's learning capacity using a consistent test benchmark that we created for fair evaluations of the generalization capabilities of the models. The size of our test benchmark, comprising 12,000 pairs, is much larger in sample size than the training datasets. Additionally, two subsets of the synthetic dataset OpenFWI \cite{NEURIPS2022_27d3ef26}, which is openly available, are employed to assess the efficacy of the deep model we constructed. These subsets are known as CurveVel-B and CurveFault-A.

Our contribution to the existing literature can be summarized as follows:

$\bullet $ We propose a novel end-to-end CNN-based densely connected encoder-decoder neural network architecture (SVInvNet) specifically designed for seismic velocity inversion task. SVInvNet is distinguished by its significantly reduced parameter count, thereby enhancing computational efficiency while maintaining high performance levels.

$\bullet $ We present two distinct large scale datasets, differentiated by the presence and absence of noise. In this context 18,000 paired seismic data and velocity models and 18,000 noisy pairs are prepared and used in this research. We will make this dataset publicly available soon.

$\bullet $ We establish a comprehensive benchmarking procedure for precise assessment and comparison of the trained deep learning (DL) models. Our approach uses a benchmark test set considerably larger than the training dataset, enhancing the evaluation of the model's generalization capabilities in diverse scenarios.

$\bullet $ We provide a comprehensive analysis of the impact of varying training dataset sizes on the learning process for the seismic velocity inversion problem. 

The remainder of this paper is organized as follows: Section \ref{sec:RW} covers the Related Work, providing a background and context for our study by reviewing relevant literature and previous studies in the field. In Section \ref{sec:MI}, we present our Methodology and Implementation details. This section is further divided into subsections, where we discuss the preparation of the dataset, describe the velocity models and seismic data, and elaborate on the proposed deep model architecture, including specifics of the dense block, encoder and decoder networks. Following this, Section \ref{sec:ER} is dedicated to Experiments and Results, where we analyze the performance of our proposed method, presenting comprehensive experimental findings and interpretations. Finally, we conclude our work in Section \ref{sec:Con}.

\section{Related Work}
\label{sec:RW}
Numerous studies across various disciplines have demonstrated the effectiveness of CNNs in extracting features from an input and converting it into a different format or dimension \cite{Minaee}. This technique has been applied in various areas of exploration seismic, including seismic wave propagation simulation \cite{moseley}, geological unit classification \cite{Hall}, reservoir study \cite{Lim}, low-frequency component recovery \cite{Ovcharenko}, and random noise suppression \cite{Saad}. 

For the first time in the field of seismic inversion, Roethe and Tarantola\cite{Rote} were able to estimate a one-dimensional velocity model from seismic data by training neural networks. In their study, the arrival time of the reflected seismic wave at the receivers were used as input data and the depth and velocity values of the single layer velocity model as the target. In their paper, Tarantola, a pioneer of the FWI method, and his colleague suggested that neural networks could serve as an alternative to traditional inverse methods.

 However, due to the limited computational facilities and insufficient development of the neural network method, this method has been ignored in seismic studies for many years. Recently, as advancements in computer technology and machine learning algorithms continue, especially in DL methodologies, we are witnessing a growing trend of deep network applications in the seismic exploration domain, consistent with developments observed in other scientific fields. In the domain of seismic inversion, you can find a timely overview of DL approaches up to 2020 in Adler et al. \cite{Adler-workflows}.
 
Lewis et al. \cite{Lewis} created a good initial velocity model for the FWI method to visualize the region under the salt dome using DL. For the training of the salt dome structure, they used real data (boreholes, well logs, and seismic data collected over many years) containing the salt dome structure in the hydrocarbon exploration region in the Gulf of Mexico. In this local study, post-stacking seismic gathers were used as the input.

In their 2018 study, Araya-Polo et al. \cite{Araya-Polo}. employed a deep neural network, GeoDNN, to predict 2D synthetic velocity models. They generated artificial seismic gathers using the acoustic seismic wave equation within these 2D synthetic models. For each model, a common midpoint (CMP) velocity cube was created through traditional velocity analysis. This cube is then utilized as the input, with the velocity model being the target during training. GeoDNN approximately identified the number of layers in stratified velocity models. However, the predicted interfaces are smoother than the actual velocity contrasts. For models with fault or salt body, the faults appeared less distinct and the salt bodies  are predicted with lower velocity value compared to the ground truth and do not fully correspond to the desired salt bodies in shape.

 Wu et al. \cite{Wu} developed InversionNet using an encoder-decoder architecture and implemented a comprehensive end-to-end training process. In contrast to Araya-Polo et al. \cite{Araya-Polo} who incorporated velocity analysis, Wu and his team directly employed seismic shot gathers in their approach. InversionNet was able to predict velocity models with four-layer, horizontal interfaces, and fault with certain success. They use 50,000 pairs of seismic shot gathers and velocity models as training dataset and 10,000 pairs for testing. Seismic shot gathers are calculated for 3 sources and 32 receivers which are evenly distributed along the top boundary of the velocity models.

Yang and Ma \cite{Yang} presented seismic shot gathers as input to the deep model to train velocity models with smooth interface curvatures containing salt body using the UNet design. They employed 1600 synthetic velocity models for the pretraining of deep model. Subsequently, the SEG dataset was utilized to extract 130 velocity modes, and establish the transfer learning \cite{PanYang}. The dimensions of their velocity models is 201$\times$301 grids. For the computation of seismic shot gathers, the researchers utilized 29 sources and 301 receivers. In this study, the trained deep model demonstrates a superior capability in predicting velocity models incorporating a salt body with arbitrary shape and position, surpassing the performance of the FWI method. They used seismic shot gathers without any preprocessing.

Li et al. \cite{Li} introduced SeisInvNet, a new deep neural network design. This network predicted velocity models with curved (wavy) interfaces. However, their models did not feature fault and salt dome structures. They categorized their velocity models into four groups, ranging from 2 to 5 layers, with 3,000 models in each group. For their training dataset, they used 2,750 models from each group, totaling 11,000 models, alongside 20 shot gathers with 32 receivers each. They employed 1000 models for testing.

 Liu et al. \cite{Liu} enhanced the input component of SeisInvNet to train velocity models with curved interfaces, encompassing geological features like faults and salt domes. In addition to shot gathers, they incorporated receiver gathers as input. Their velocity models span five categories (5-9 layers), with each category further classified into three types: dense layer, faulty, and those containing salt body. For every velocity model, they generated 20 shot gathers using 32 receivers. Their training dataset comprises 1,100 models for each type, totaling 16,500 models; they employed 1500 models for testing. \\
Research continues to advance in the pursuit of more reliable results. In the aforementioned studies, the estimation of the seismic velocity model has been approached as a regression problem. However, Simon et al. \cite{SIMON2024105534} employed both regression and classification methods to estimate the velocity model in the time domain in a comparative study. According to the results presented in their study, the classification method demonstrates a lower error rate in estimating velocity models compared to the regression method.

\section{Methodology and Implementation}
\label{sec:MI}
\begin{figure}[!t]
\centering
\includegraphics[width=0.47\textwidth]{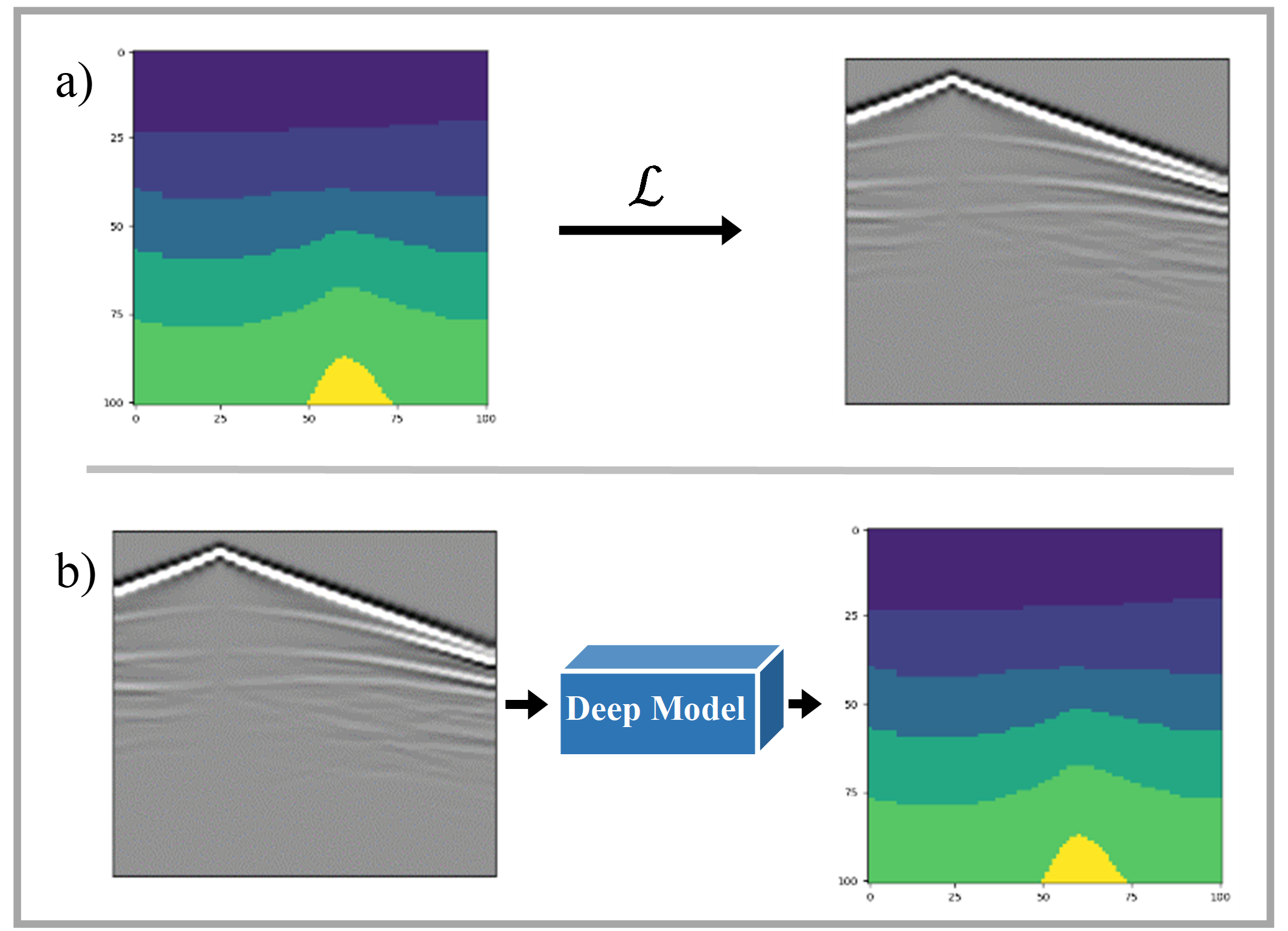}
\caption{Diagram illustrating (a) forward modeling and (b) inversion processes in seismic velocity studies utilizing DL-based models. The seismic propagation equation, $\mathcal{L}$, facilitates the computation of seismic data; however, no explicit equation is available for deriving velocity based on existing seismic data.}
\label{fig1}
\end{figure}

Using the seismic wavefield propagation equation the seismic data can be calculated in a velocity model. As shown in Fig. \ref{fig1}(a) we can write this process with an operator such as $\mathcal{L}$.

\begin{equation}
\label{eq:D}
D = \mathcal{L}(m_v)
\end{equation}

In (\ref{eq:D}), $D$ is seismic data, and $m_v$ is velocity model parameters.
In this study the $\mathcal{L}$ is 2D acoustic wave equation (see Eq. \ref{eq:wave}). It can be solved by different numerical methods but there is no equation or straight way to reverse the direction of $\mathcal{L}$ in Fig. \ref{fig1}(a). Mathematically the inverse of $\mathcal{L}$ gives $m_v$ as shown in Eq. (\ref{eq:mv}).

\begin{equation}
\label{eq:mv}
m_v = \mathcal{L}^{-1}(D)
\end{equation}

The inverse operator (represented as  $\mathcal{L}^{-1}$) is crucial for predicting the velocity model parameters that best fit the empirical seismic data. This process represents a complex nonlinear inversion problem, where adjustments to the velocity parameters do not manifest in a directly proportional response within the seismic data observations. Furthermore, the issue is inherently ill-posed, as it permits the existence of several sets of velocity parameters that can produce equivalent seismic data, thereby complicating the solution's uniqueness and stability.

Addressing this nonlinear ill-posed problem, researchers have explored a diverse range of methodologies. These span from global methods, which utilize Eq.(\ref{eq:D}) to identify an optimal $m_v$ by investigation in velocity model domain \cite{SenStoffa}, to derivative-based approaches. The latter initiate with an initial model $\bar m_v$, subsequently compute the corresponding seismic data $\bar D$, and adjust the $\bar m_v$ parameters to minimize the local gradient of $L(D- \bar D)$ \cite{Lailly}, \cite{Taran84} (L is the loss function). More recent advancements have seen the emergence of DL-based techniques, providing innovative solutions to this long-standing problem \cite{Araya-Polo}, \cite{Wu}, \cite{Yang}, \cite{Li}, \cite{Liu}.

The DL model can be mathematically represented as shown in Eq. (\ref{eq:f}):

\begin{equation}
\label{eq:f}
\mathcal{F}(D^{n},\mathcal{W}) \rightarrow m^{n}_v
\end{equation}

Here, $\mathcal{F}$ denotes the DL function responsible for transforming the seismic data, $D^n$, into a velocity model, $m^{n}_v$. $\mathcal{W}$ represents the set of learnable parameters or weights within the model. If N is the total number of training dataset, then each individual training sample can be denoted by n, such that n $\in$ N. The function $\mathcal{F}$ is composed of multiple neural network layers, each with varying depth and connection patterns, extending from the input to the output layer. The input data propagates through these layers to produce an output. This output, in supervised method, is then compared to the ground truth or target and the distinction between them is quantified through the utilization of the loss function. The success rate of the deep model is influenced by both the quantity and comprehensiveness of the training dataset.

In this study, seismic sources were evenly distributed above the velocity models and scanned at various angles to produce seismic shot gathers. A total of 20 shot points were utilized, resulting in 20 distinct seismic shot gathers. These shot gathers are presented to the deep neural networks as input, with the corresponding velocity model serving as the ground truth. 

In our research, we employed Convolutional Neural Networks (CNNs) based deep model to establish a clear link between the seismic data's characteristics, such as direct arrivals, reflections, scattering, and potential refractions, and the underlying velocity model parameters (Fig. \ref{fig1}(b)) such as inter-layer velocity values, number of layers, layers interface shape, salt dome, and fault properties. To achieve this, we designed a neural network with a layered structure that is deep enough to maintain the integrity of the gradient flow during training. We introduced an encoder-decoder framework in our neural network, which allows us to train on complex models directly, bypassing the need for initial data manipulation or pre-processing. Furthermore, by integrating a Dense network structure, we were able to construct more robust neural network layers while ensuring that the gradient signal does not weaken significantly as it travels through the network. The details of this neural network design are elaborated in Section \ref{subsec:Architecture}.

\subsection{Dataset Preparation}
To establish the training and test datasets, we first generated velocity models, followed by seismic shot gathers. In all, we prepared 18,000 velocity models. These models are categorized into five groups, with the number of layers ranging from 4 to 8. Each group is further divided into three subgroups: stratified, faulty, and salt dome, with each subgroup containing 1,200 models. For every velocity model, we calculated 20 shot gathers. Each seismic data has a time dimension of 1 second and encompasses 34 receivers. All computations are carried out in a constant density acoustic environment, with velocity values increasing in correlation with depth.

\subsubsection{Velocity Models}
In order to prepare realistic velocity models, we closely examined published stratigraphic examples and incorporated geological principles into our approach. Consequently, inclined, undulating interfaces, normal and reverse faults with varying angles and fault throw, and salt bodies with arbitrary size, shape and horizontal positions are used. 

To prepare interfaces resembling subsurface stratigraphic formations, we generated 116 distinct interface curves using various mathematical functions, including trigonometric, polynomial, and logarithmic. Eq. (\ref{eq:l1}) through (\ref{eq:l3}) display three examples of these functions. 

\begin{equation}
\label{eq:l1}
l_1 = \sqrt{x}+5\log(15x+1)\sin(0.14x)\cos(0.3x+20)
\end{equation}
\vspace{0.05in}
%---
\begin{equation}
\label{eq:l2}
l_2 = 0.09x^2-3.6\sin(0.6x+2)+5(x+1.5)-0.8x
\end{equation}
%---

\begin{equation}
\label{eq:l3}
l_3 = e^{(0.1x-0.11)}-1.5\sin(x-6)+2.1\frac{x}{x+3}
\end{equation}

The interfaces are saved as integer values. The velocity model is created by randomly selecting the depth, interface shape, and velocity value for each layer, with the process commencing from the top layer and continuing downwards. This process for each velocity layer is illustrated in Alg. \ref{alg:multi-layer}.

Two principles are considered in this process. First, the velocity value increases with depth, meaning that the lower layer always has a higher velocity than the upper one. Second, to partially maintain the geological sediment deposition logic between the layers, the shape of the interfaces above is influenced by those below. Hence, novel and stochastic interface configurations emerge, resulting in an expansion of the number of interfaces employed in the structure of velocity models beyond the initially designated 116 functions. Consequently, the stochastically chosen interfaces merge with the ones below. The inter layer velocity value varies between [1500, 4000] m/s. 

\begin{algorithm}
\caption{Multi-Layer Filling Algorithm}
\label{alg:multi-layer}
\begin{algorithmic}
\REQUIRE Stochastic selection of parameters, Total number of layers \(N\)
\STATE Initialize \(d_{prev} \leftarrow 0\)
\STATE Initialize \(L_{prev} \leftarrow 0\)
\FOR{\(i = 1\) to \(N\)} % Here N is the total number of layers
    \STATE \(V_i \leftarrow\) Choose velocity value for the \(i\)-th layer
    \STATE \(d_i \leftarrow\) Choose depth of the \(i\)-th layer
    \STATE \(L_i \leftarrow\) Choose interface of the \(i\)-th layer
    \STATE \(spaceStart \leftarrow d_{prev} + L_{prev}\)
    \STATE \(spaceEnd \leftarrow d_{prev} + L_{prev} + d_i + L_i\)
    \STATE Fill the space between \(spaceStart\) and \(spaceEnd\) with \(V_i\)
    \STATE Update \(d_{prev} \leftarrow d_{prev} + d_i\)
    \STATE Update \(L_{prev} \leftarrow L_{prev} + L_i\)
\ENDFOR
\end{algorithmic}
\end{algorithm}

The minimum velocity difference between layers is 200 m/s. In a medium characterized by constant density, the velocity contrast across layer boundaries plays a crucial role in determining the reflection coefficient. Particularly, within layers with higher velocity values, a small contrast results in a significant decrease in the amplitude of the reflected wave. The size of the velocity models is 100$\times$100 grid points.

We created fault models by incorporating fault line code into the prepared layered models algorithms. The angle of line, fault type (normal and reverse), and fault throw value is chosen stochastically. According to the type of fault the hanging wall part of the model moved up or downward across the fault line. 

\begin{figure}[!t]
\centering
\includegraphics[width=0.45\textwidth]{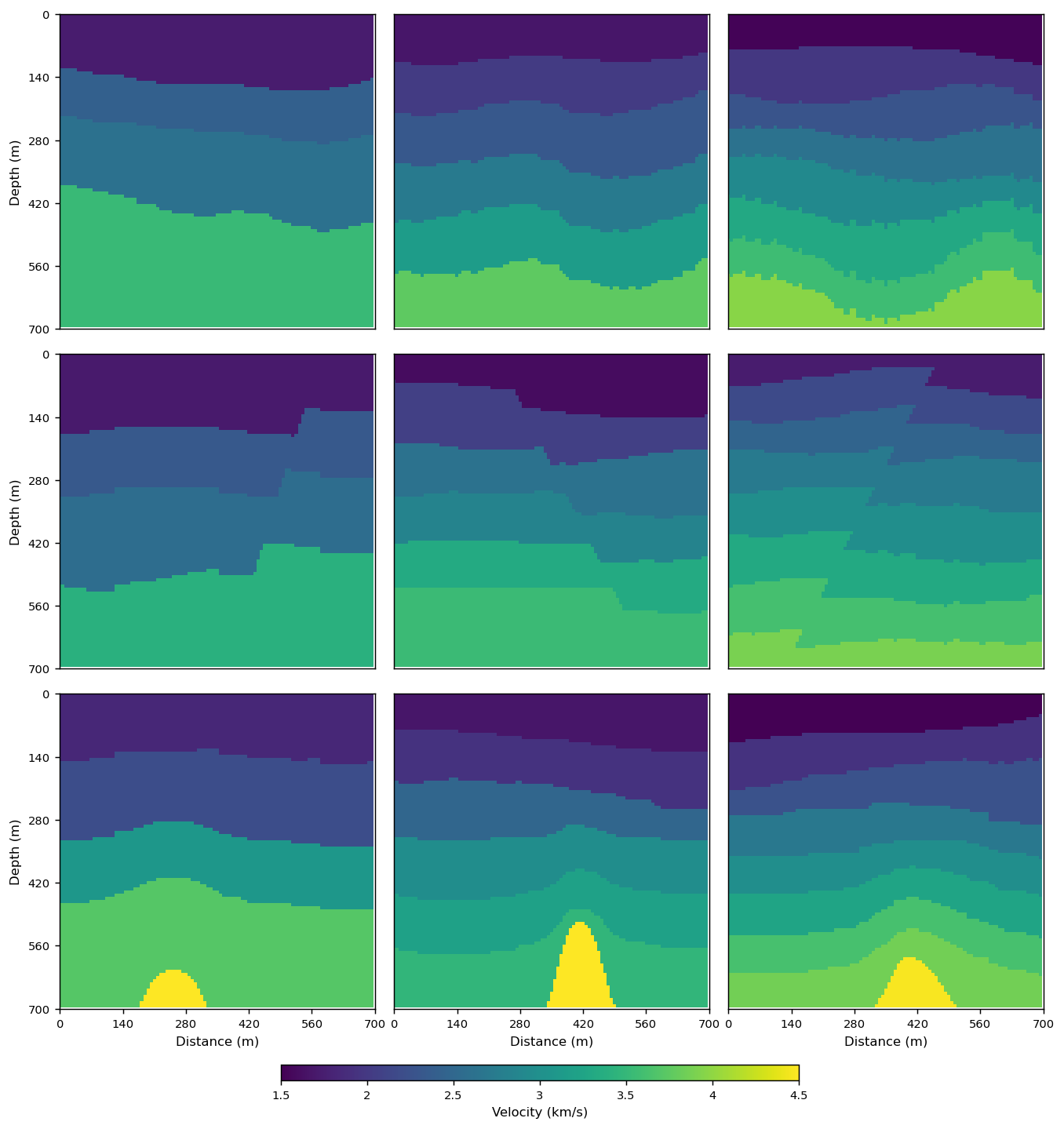}
\caption{Samples from our dataset; top: layered, middle: fault, bottom: salt dome velocity model samples. The samples are selected from 4, 6, and 8 layered models. The model's spatial resolution is $700\times700$ m.}
\label{fig2}
\end{figure}

To construct the models with a salt dome at least four different Gaussian functions are generated stochastically and  their combinations are utilized. This approach enables the creation of salt domes with diverse sizes and shapes. The dome's velocity values change between [4350, 4550] m/s. The intrusion of a salt dome into the overlying sediments deform the shape of the upper layers. For inducing this deformation in the upper layers of the salt domes, we employed wider Gaussian functions. Some sample velocity models are depicted in Fig. \ref{fig2}.

\subsubsection{Seismic Data}
In order to calculate the seismic shot gathers presented to the deep model as input, 34 receivers and 20 sources are distributed uniformly along the upper boundary of the velocity models. To capture data from the lateral portion of the velocity models, we positioned the receiver at both the initial and final grid points. Given a receiver interval of 3 grids with an evenly distributed setup, the total number of receivers is determined to be 34. The source locations are initiated from the third grid point and proceeded with intervals of 5 grids, reaching the 98th grid. The time progression of the seismic wavefield in a 2D acoustic environment is calculated by Eq. (\ref{eq:wave}).

\begin{equation}
\label{eq:wave}
\frac{\partial^2P}{\partial t^2}-v^2(\frac{\partial^2P}{\partial x^2}+\frac{\partial^2P}{\partial z^2}) = \mathcal{S}(t,x,z) 
\end{equation}

\begin{figure}[!t]
\centering
\includegraphics[width=0.48\textwidth]{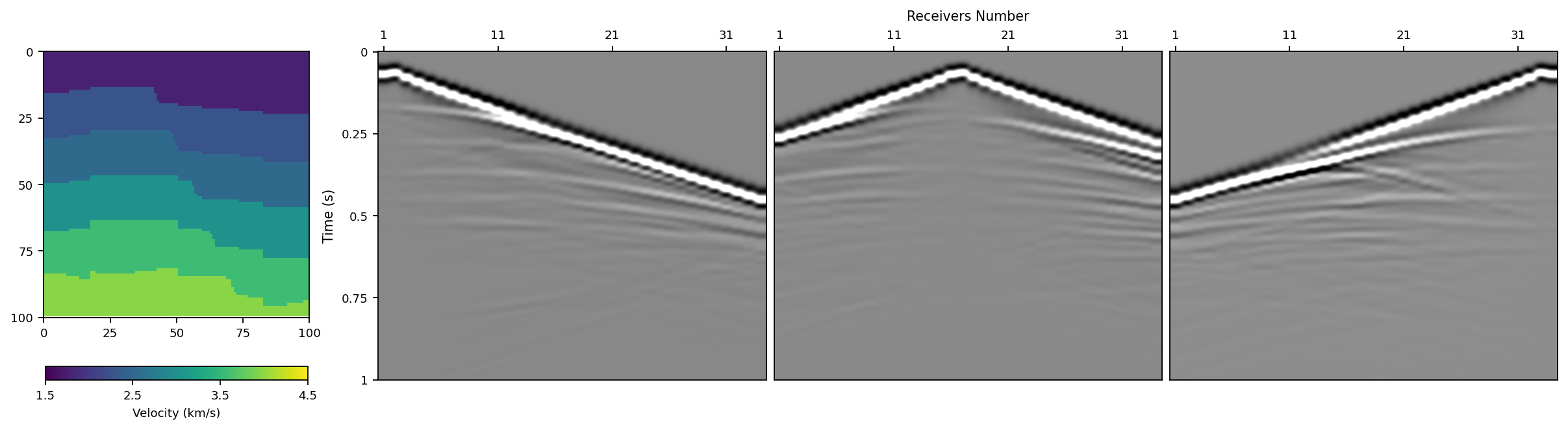}
\caption{The velocity model and its associated synthetic shot gathers at the first, tenth, and twentieth source locations respectively, from left to right.}
\label{fig3}
\end{figure}

Where P is the acoustic wavefield, $v$ is the seismic velocity of the medium, $x$ and $z$ are spatial coordinates, $\mathcal{S}$ is the source function and $t$ is time. The numerical solution of Eq. (\ref{eq:wave}) is implemented in the time domain by 2D finite difference method \cite{kellysynthetic}. Seismic data is computed for a duration of 1 second at each individual receiver location. The Ricker wavelet with a dominant frequency of 20 Hz is used as the seismic source. To mitigate dispersion in the numerical computation, the spatial environment is sampled at intervals of 7 meters, adhering to the Alford et al. \cite{Alford} condition and considering $v_{min}=1500$ m/s alongside the source frequency (20 Hz), in other words $\Delta x=7$m. To simplify the calculation $\Delta z= \Delta x$. Considering the $\Delta x$ value and $v_{max}=4550$ m/s, to ensure the accuracy of numerical computations, we set the time sampling intervals to 1ms (Lines et al.\cite{Lines} condition), in total 1000 time samples. In order to minimize the effect of boundaries, 20 grid points are added on the right, left and bottom of each velocity model and Clayton and Engquist \cite{Clayton} absorbing boundary conditions are used. Consequently, seismic data of dimensions [20, 1000, 34] are acquired for each velocity model. Fig. \ref{fig3} illustrates 1th, 10th, and 20th shot gathers corresponding to the randomly chosen velocity model.

To partially reduce the disparity between the synthetic and real seismic data, both coherent and random noise are added to the prepared dataset. Coherent noise is derived from the characteristics of surface waves. Hence, spike patterns, characterized by velocity values ranging from 250 to 450 m/s, are generated in such a way that the maximum amplitude is observed at the proximal receiver. These patterns exhibited a progressive attenuation with temporal advancement. Subsequently, a Ricker wavelet with a central frequency randomly chosen between 8-17 Hz as the source of the surface waves is convolved with the spikes gather. Through the application of this stochastic algorithm to each velocity model, 18,000 coherent noise gathers are produced, demonstrating the impact of surface waves.

\begin{figure}[!h]
\centering
\includegraphics[width=0.48\textwidth]{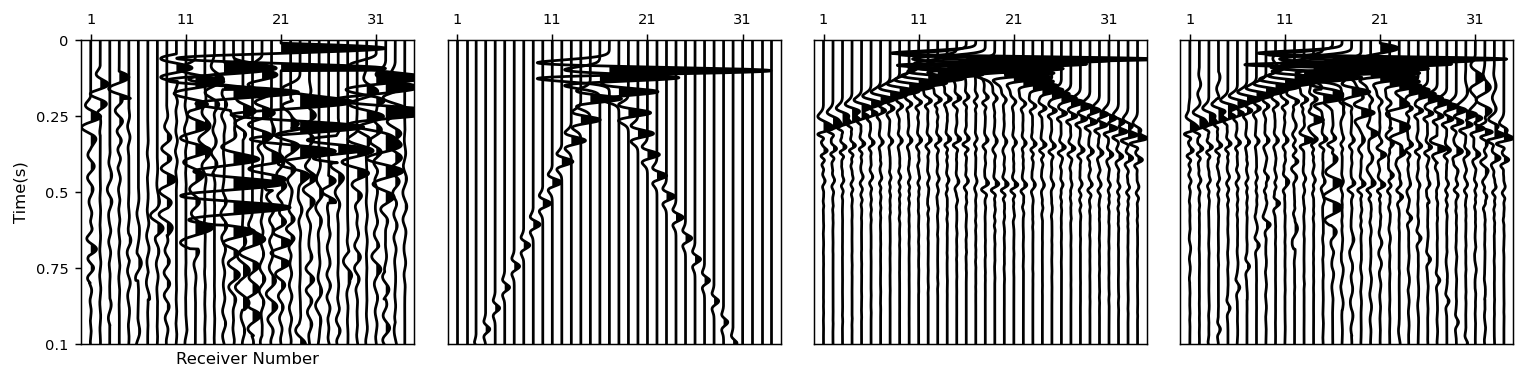}
\caption{The samples of stochastic, coherent, noiseless and noisy gather are arranged from left to right, respectively. The gathers are plotted on the same scale.}
\label{fig4}
\end{figure}

To generate random noises, initially, white noise with a randomly selected standard distribution, and zero mean is created. In the subsequent phase, certain randomly chosen segments of the produced noise are assigned a value of zero, and it is convolved with a sine function characterized by a randomly selected frequency value falling within the range of 13 to 17 Hz. The computed noisy data is then added to the seismic shot gathers, resulting in the acquisition of a noisy dataset. One sample of the stochastic, coherent, noiseless and noisy shot gather is shown in Fig. \ref{fig4}.

\subsection{Architecture of the Deep Model}
\label{subsec:Architecture}
In this study, we take the design of the InversionNet \cite{Wu} deep model as the baseline. Similar to InversionNet, we employ one-dimensional (1D) CNN kernels to reduce the time axis of the input data to the number of receivers within the initial four layers. However, unlike InversionNet, we do not reduce the size of the features below 6$\times$6. This decision is based on the understanding that the amplitude and the arrival times of adjacent traces contains crucial information pertaining to the structure of the velocity model interfaces. Various trials were conducted to determine the smallest size of the features in our deep model, with dimensions \(4 \times 4\), \(5 \times 5\), \(6 \times 6\), \(7 \times 7\), and \(8 \times 8\) being evaluated. Following the analysis of the results, dimension \(6 \times 6\) was chosen as it provided the best performance.

In our deep model architecture (Fig. \ref{fig6}), both the encoder part and the decoder part consist of different dense blocks (depicted as blue blocks in the figure) with fixed feature size and transition layers in order to speed up the information flow and prevent the fading of the gradients during backpropagation. Furthermore, certain outputs from the dense blocks in the encoder segment are concatenated with features of the same size in the decoder part. To avoid excessive layer addition, the experimentation commenced with a shallow CNN model. Based on the findings, subsequent improvements are implemented by incorporating new layers and dense blocks into the model structure. Owing to the structured design of the dense blocks in our architecture, we can expand the number of layers and concatenate their output features without the need for a progressive increase in channel size.

\subsubsection{Dense Block Structure}
\begin{figure}[!h]
\centering
\includegraphics[width=0.48\textwidth]{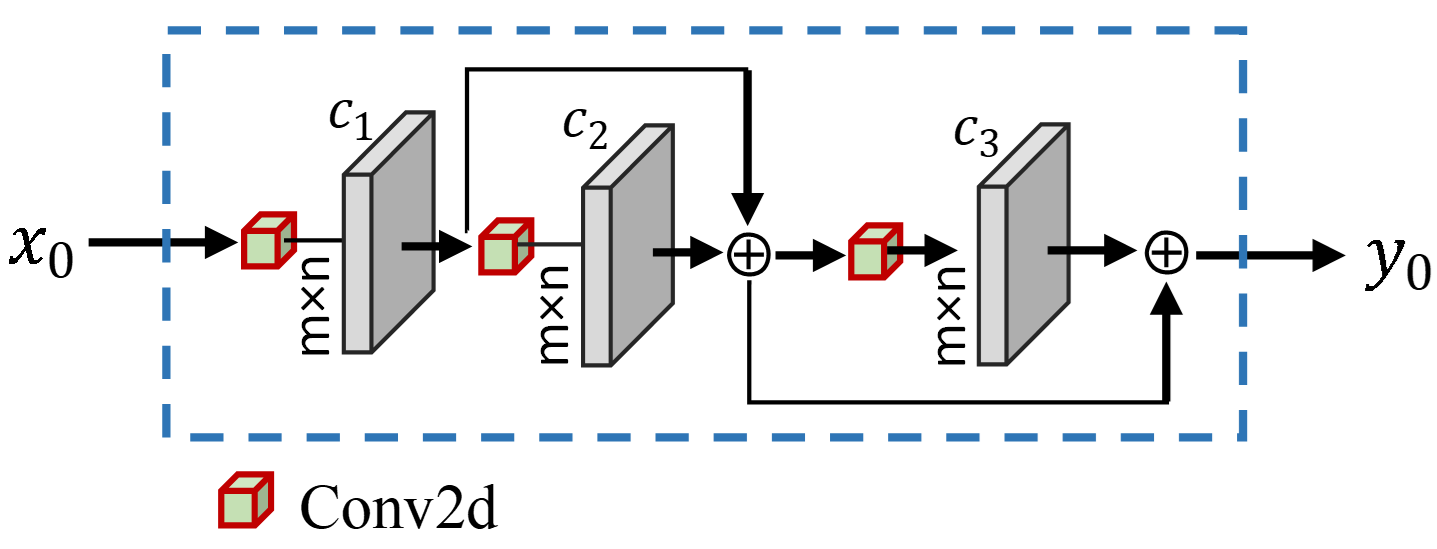}
\caption{A dense block with 3 layers.}
\label{fig5}
\end{figure}

Within each dense block, there are 3 CNN layers, with the feature size matching that of the initial layer. Fig. \ref{fig5} illustrates the schematic representation of the i-th dense block, and the connections between its layers. Equations (\ref{eq:d1})-(\ref{eq:d5}) describe i-th dense block mathematically. 

\begin{equation}
\label{eq:d1}
x^{i}_1 = L^{i}_1(x_0)\hspace{0.08in};\hspace{0.08in} C_{out}=c_1
\end{equation}
\vspace{0.05in}
%---
\begin{equation}
\label{eq:d2}
x^{i}_2 = L^{i}_2(x_1)\hspace{0.08in};\hspace{0.08in} C_{out}=c_2
\end{equation}
\vspace{0.05in}
%---
\begin{equation}
\label{eq:d3}
x^i_{1,2}=concatenate(x^i_1,x^i_2)
\end{equation}
\vspace{0.05in}
%---
\begin{equation}
\label{eq:d4}
x^{i}_3 = L^{i}_3(x^i_{1,2})\hspace{0.08in};\hspace{0.08in} C_{out}=c_3
\end{equation}
\vspace{0.05in}
%---
\begin{equation}
\label{eq:d5}
y_0 = concatenate(x^i_1,x^i_2,x^i_3)
\end{equation}

\begin{figure}[!t]
\centering
\includegraphics[width=0.49\textwidth]{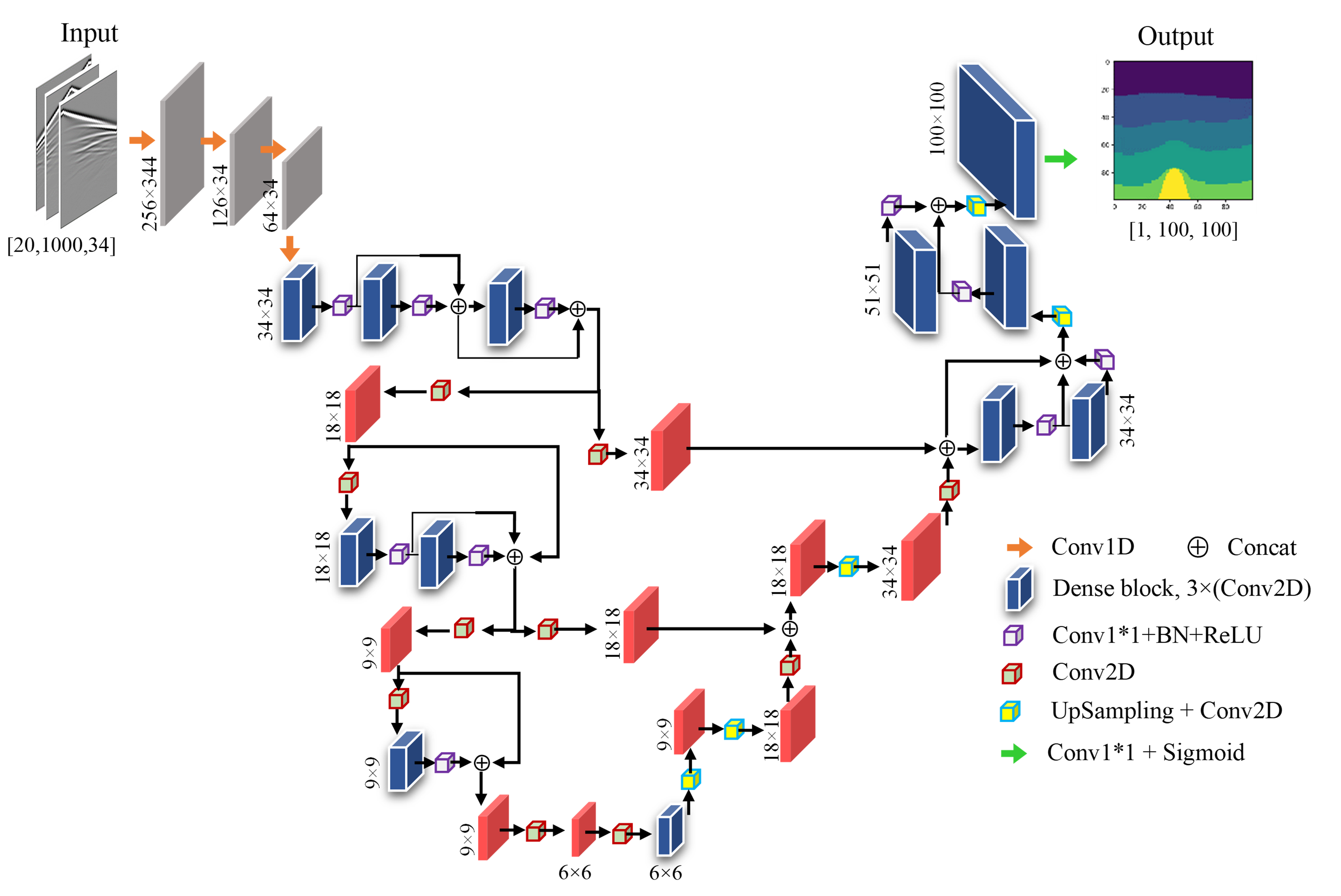}
\caption{Proposed SVInvNet architecture for seismic velocity inversion. Best viewed in the electronic version with zoom.}
\label{fig6}
\end{figure}

Here, $L^i_j$ is the j-th layer of a dense block i, $x_0$ is input to the dense block and $y_0$ is the output of i-th dense block; $x^i_j$ is the output of the j-th layer. $c_j$ shows the channel size (i.e. number of feature maps generated), which determines the depth of the output volume, in each layer. In our architecture $c_j$= 64 and $y_0$ has 192 feature maps, and $L_j$ contain a 3$\times$3 convolution (Conv), batch normalization (BN) \cite{Ioffe} layers, and rectified linear unit (ReLU)\cite{GlorotReLU} activation function. For the sake of convenience, this combination, conv2d+BN+ReLU, will be referred to as the Conv2d in our discussion from this on. In the transition layers between the dense blocks with identical feature sizes the channel size of the output is typically reduced (as shown with purple cubes in Fig. \ref{fig6}). As evident from the design of the dense blocks, the extracted features are concatenated and used in the subsequent layers. Consequently, the input reaching these layers is richer in information. Additionally, the derivative information reaching the output part ($y_0$) of the dense block through backpropagation can be instantly distributed to all inner layers.

\subsubsection{Encoder Network}
Through the utilization of 1D-CNNs for the initial four layers (depicted with orange arrows in Fig.6), the time dimension of the input data is reduced based on the number of receivers. In other words, the output dimension of the fourth layer is 34$\times$34. Subsequently, the data dimensions are progressively reduced to 18$\times$18, 9$\times$9, and 6$\times$6 through the application of 2D-CNNs. For the 34$\times$34 dimension, there are three dense blocks, whereas for the other dimensions, there exist 2, 1, and 1 dense blocks, respectively. There are three Conv2d layers within each dense block. Each layer's input comprises the outputs of the preceding layers. Therefore, the CNN kernels are able to utilize the features extracted in the preceding layers to access low level features.

The outputs of the dense blocks with identical feature sizes are concatenated and subsequently fed into the Conv2d layer where their size can be reduced via stride values exceeding one. As a result, the size of output channels decreases during the transition from these layers and the output from these layers is used as the input to a new dense block. In Fig. \ref{fig6}, these layers are depicted using small red-green cubes.

\subsubsection{Decoder Network}
The features extracted within the layers of the encoder model are progressively scaled to dimensions of 9$\times$9, 18$\times$18, 34$\times$34, 51$\times$51, and 100$\times$100 in order to construct the output corresponding to the velocity model. For sizes 34$\times$34 and 51$\times$51, two dense blocks are utilized for each, whereas a single dense block is employed for size 100$\times$100. The process of scaling up takes place within the layers, which comprise upsampling and Conv2D operations. (The small blue-yellow cubes in Fig.\ref{fig6} depict these layers.) In order to facilitate the information transfer process and consolidate the decoder part features, the outputs of the dense blocks with a size of 18$\times$18 and 34$\times$34 in the encoder part are also concatenated with their corresponding counterparts in the decoder part. Due to the incorporation of dense blocks, this architecture enables an increase in the channel size, allowing for the processing of larger feature sizes without a rise in the number of parameters. The inclusion of a 100$\times$100 dimensional dense block before the output layer proves to be instrumental in the efficacy of the deep model. The output of this dense block undergoes a Conv$1\times1$ + Sigmoid activation function, yielding the deep model output (depicted with a green arrow in Fig.6). 

In this proposed architecture, on the one hand, features are extracted from seismic data and aligned with the size and dimensions of the velocity model, while on the other hand, the essential information required for this alignment can be easily transferred between the input and ground truth. The number of parameters of SVInvNet is about 4 million. 

The output of this architecture is compared with the ground truth, and the discrepancy between them is computed using the loss function. The loss function should be capable of capturing the disparity among the velocity values as well as the shapes of the interfaces and geological structures between two velocity models. Hence, in this section of the architecture, we employ both mean absolute error (L1) and structural similarity index measure (SSIM) as loss functions, and we assign weights to each of them using two coefficients ($\lambda_{L1}$ and $\lambda_{SSIM}$ ) to enhance the derivatives of the loss functions.

\newcommand{\RomanNumeralCaps}[1]
    {\MakeUppercase{\romannumeral #1}}

\section{Experiments and Results}
\label{sec:ER}
Given that the number of layers in each velocity model ranges from 4 to 8, the total number of models, including layered, faulted, and salt dome types, is calculated as 5$\times$3$\times$1200. From each of the 15 different subgroups of data, 800 models are randomly selected as part of the test benchmark. Hence, the total number of test dataset size is 12,000. The remaining models in the dataset (i.e. 6,000 models) are used as the training dataset. To investigate the impacts of varying the quantity of training data on the mapping or seismic inversion process, we prepared five different training datasets. Accordingly, we selected various quantities of subgroups, as 50, 100, 200, 300, and 400 models from each type. Consequently, we undertook five separate training processes with 750 (15$\times$50), 1,500, 3,000, 4,500, and 6,000 training dataset sizes and evaluated the trained deep models using the same test benchmark that we prepared using 12,000 models. For the purposes of clarity and convenience, we introduce a nomenclature to denote the various training datasets. Specifically, we assign Roman numerals I-V to represent training datasets corresponding to sample sizes respectively. Hence, these datasets will be referred to as TD-\RomanNumeralCaps{1}, TD-\RomanNumeralCaps{2}, TD-\RomanNumeralCaps{3}, TD-\RomanNumeralCaps{4}, and TD-\RomanNumeralCaps{5}, respectively. The training datasets and related information for each are provided in Table \ref{table:TrainDataset}.

The number of test data is at least twice that of the training datasets, enabling a more accurate investigation into the generalization ability of the trained deep models\cite{goodfellow2016deep}. For example, TD-\RomanNumeralCaps{4} contains 300 models of each velocity subgroups, in terms of number of samples, it corresponds to 37.5\% of the test benchmark size. The training conditions remained identical across all five training processes. Employing a consistent nomenclature for the trained models, we extend the use of Roman numerals I-V to the designation of our train models. For instance Model-\RomanNumeralCaps{1} will be corresponding to the model that is trained with TD-\RomanNumeralCaps{1} and so on.

 We established the network hyperparameters based on the results of various experiments. For optimization, the Adam optimizer \cite{kingma2014adam} is utilized with a batch size of 32, and the initial learning rate is set to 5e-3. During the training, the learning rate is adjusted by gradually decreasing it in accordance with the optimization process. The number of epochs for all training processes is set to 500.

To compute the misfit and similarity between deep model predictions and the ground truths in the test stage, we utilize mean absolute error (L1) and mean square error (L2), structural similarity index measure (SSIM), and multi-structural similarity index measure (MSSIM). A successful estimation of velocity model or velocity inversion is characterized by lower L1 and L2 values, as well as higher SSIM and MSSIM values. These values are computed for the normalized dataset. All training and testing procedures are executed on GPUs, on an NVIDIA RTX A4500 graphics card, using PyTorch library.

%---- Table
\begin{table} [!h]
\begin{center}
\caption{The training datasets and their respective subgroups}
\label{table:TrainDataset}
\begin{tabular}{c c c c c c}
\hline 
Training dataset & \multicolumn{1}{|c}{TD-\RomanNumeralCaps{1}} & TD-\RomanNumeralCaps{2} & TD-\RomanNumeralCaps{3} & TD-\RomanNumeralCaps{4} & TD-\RomanNumeralCaps{5}\\
\hline \hline
\multicolumn{1}{c|}{Sample size} & 750 & 1500 & 3000 & 4500 & 6000 \\
\hline
\multicolumn{1}{c|}{Layer counts} & \multicolumn{5}{c}{Ranges uniformly from 4 to 8 layers}  \\
\hline
\multicolumn{1}{c|}{Structural features} & \multicolumn{5}{c}{Dense layered - Faulty - Salt dom} \\
\hline
\multicolumn{1}{c|}{Seismic data} & \multicolumn{5}{c}{34 receivers, 20 sources, and 1s records}  \\
\hline
\multicolumn{1}{c|}{Noise content } & \multicolumn{5}{c}{two datasets: noisy and noiseless } \\
\hline
\end{tabular}
\end{center}
\end{table}
%-----

 \subsection{Baseline Model}
To compare the performance of SVInvNet, we  employed the InversionNet model, proposed by Wu et al. \cite{Wu}, as the baseline, with certain modifications made to its architecture. As for the modifications, we replaced the MaxPooling function with a Convolution function in the fourth layer of the InversionNet and decreased the feature size until it reached 4$\times$4 instead of 1$\times$1. Furthermore, by substituting each deconvolution layer with an upsample + convolution approach in the decoder segment of the model, the parameter count decreased to 20 million, leading to an enhancement in the prediction capabilities.

\subsection{Results and Discussion}

Utilizing the noiseless and noisy versions of our TD-\RomanNumeralCaps{1} to TD-\RomanNumeralCaps{5}, we independently trained SVInvNet and subsequently evaluated each of the resultant five trained deep models (Model-\RomanNumeralCaps{1} to Model-\RomanNumeralCaps{5}) using our test benchmark. The misfit and similarity values of the test phases for the noiseless and noisy  dataset are shown in the Table \ref{table:ourmodels}. As anticipated, increasing the volume of the trained dataset corresponded to a decrease in the values of L1 and L2, alongside an increase in the values of SSIM and MSSIM.

\begin{figure}[!t]
\centering
\includegraphics[width=0.47\textwidth]{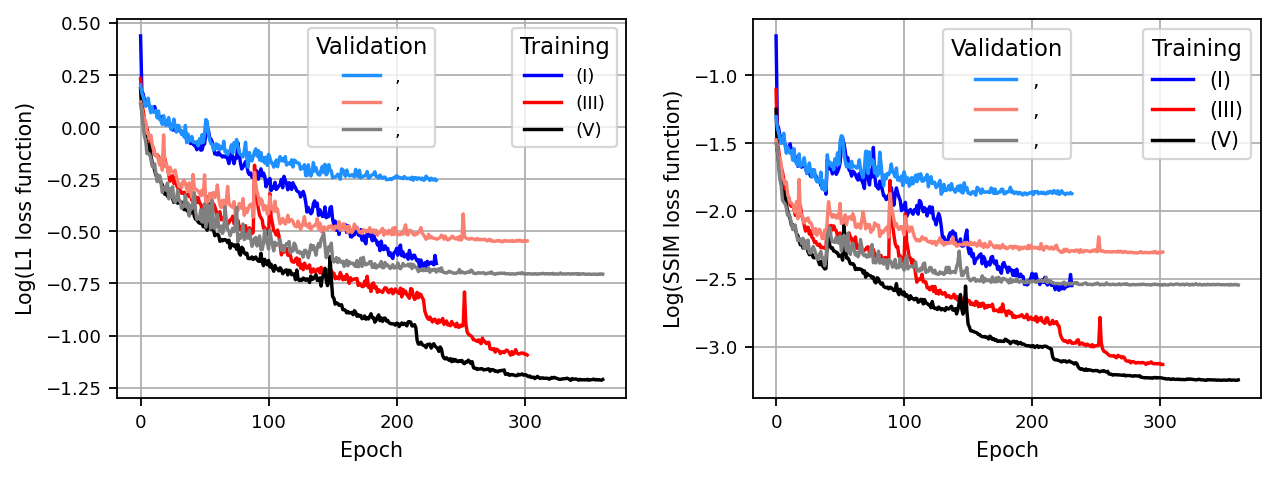}
\caption{ The L1 (Left) and SSIM (Right) loss curves on the training and validation datasets obtained during the training of Model-\RomanNumeralCaps{1}-\RomanNumeralCaps{3}-\RomanNumeralCaps{5}. Best viewed in the electronic version with zoom.}
\label{fig-loss}
\end{figure}

The L1 and SSIM loss function curves on the training and validation datasets during the training are depicted for three selected models aggregated in Fig. \ref{fig-loss}. The loss curves of the other models are bounded by Model-\RomanNumeralCaps{1} and Model-\RomanNumeralCaps{5} curves. From the plots, it is evident that Model-\RomanNumeralCaps{1} exhibits overfitting behavior. However, as the training dataset size increases (refer to Model-\RomanNumeralCaps{5}), this issue is significantly reduced. 

%----- figure 7
%------

InversionNet is trained using both the noiseless and noisy TD-\RomanNumeralCaps{5} datasets under conditions equivalent to those of SVInvNet. The statistics of misfit and similarity metrics in the test stage are presented in the Table \ref{table:ourmodels} as InversionNet-V.

Based on the results presented in Table \ref{table:ourmodels} it is evident that the performance of our proposed deep model surpasses that of the InversionNet. Even the loss values of SVInvNet, trained with TD-\RomanNumeralCaps{3}, indicate a superior performance compared to the values of the InversionNet, trained with TD-\RomanNumeralCaps{5}.

\begin{table*} [!b]
\begin{center}
\caption{The Performance of SVInvNet on the Test Set. Models I-V are trained utilizing Noiseless and Noisy TD \RomanNumeralCaps{1}-\RomanNumeralCaps{5}, respectively. The Last Column Represents Results of InversionNet Trained Using Noiseless and Noisy TD-\RomanNumeralCaps{5}.}
\label{table:ourmodels}
\begin{tabular}{ c c c c c c c c c }
\hline
 & \multicolumn{4}{c}{Noiseless Dataset} & \multicolumn{4}{c}{Noisy Dataset}  \\
\hline 
& L1 & L2 & SSIM & MSSIM & L1 & L2 & SSIM & MSSIM \\
\hline\hline
Model-\RomanNumeralCaps{1} & 0.013680 & 0.000737 & 0.9999436 & \multicolumn{1}{c|}{0.99999491} & 0.024961 & 0.001548 & 0.999856 & 0.999988  \\
Model-\RomanNumeralCaps{2} & 0.009833 & 0.000509 & 0.9999674 & \multicolumn{1}{c|}{0.99999692} & 0.018942 & 0.001044 & 0.999911 & 0.999992 \\
Model-\RomanNumeralCaps{3} & 0.006844 & 0.000356 & 0.9999810 & \multicolumn{1}{c|}{0.99999812}  & 0.014189& 0.000736 & 0.999945 & 0.999995 \\
Model-\RomanNumeralCaps{4} & 0.005301 & 0.000272 & 0.9999870 & \multicolumn{1}{c|}{0.99999870} & 0.010378 & 0.000489 & 0.999967 & 0.999997 \\
Model-\RomanNumeralCaps{5} & \textbf{0.004944} & \textbf{0.000253} & \textbf{0.9999884} & \multicolumn{1}{c|}{\textbf{0.99999882}} & \textbf{0.008034} & \textbf{0.000399} & \textbf{0.999977} & \textbf{0.999997} \\
InversionNet-V & 0.007316  & 0.000401 & 0.999978 & \multicolumn{1}{c|}{0.999997}  & 0.013142  & 0.000713 & 0.999949 & 0.999995 \\
\hline
\end{tabular}
\end{center}
\end{table*}

\begin{figure}[!t]
\centering
\includegraphics[width=0.47\textwidth]{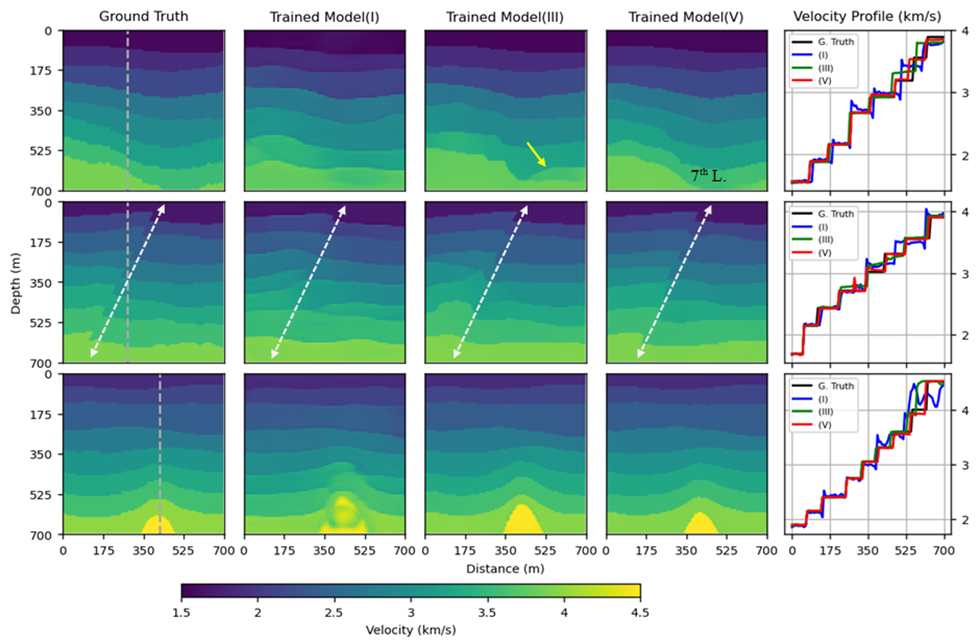}
\caption{ Ground truth velocity model and the prediction results of our Model-\RomanNumeralCaps{1} (the second column), Model-\RomanNumeralCaps{3} (the third column), and Model-\RomanNumeralCaps{5} (the fourth column). The rightmost column shows a vertical velocity profile at the gray line depicted in the ground truth model for four velocity models. Best viewed in the electronic version with zoom.}
\label{fig-Noisles-SVInvNet}
\end{figure}

Qualitative evaluations of our trained models are depicted for randomly selected test samples in Fig. \ref{fig-Noisles-SVInvNet}. Each sample is purposefully selected from 8-layer group to showcase the visuals from more challenging group with respect to the inversion for layered, fault, and salt dome categories. The ground truth and the predicted result of Model-\RomanNumeralCaps{1}, Model-\RomanNumeralCaps{3}, and Model-\RomanNumeralCaps{5} of these samples are depicted in Fig. \ref{fig-Noisles-SVInvNet}. For qualitative assessments, we pay attention to the velocity value and the interface shape of the layers, the fault line continuity, and throw value as well as the shape, position, and velocity value of the salt dome. As can be seen from the samples, the velocity values within the layers remain constant. With an increase in the size of the training dataset, the discrepancies in the approximation of this value diminish, gradually converging towards a stable and unvarying value. 

%----- figure 8
%------

%----- figure 9
\begin{figure}[!b]
\centering
\includegraphics[width=0.47\textwidth]{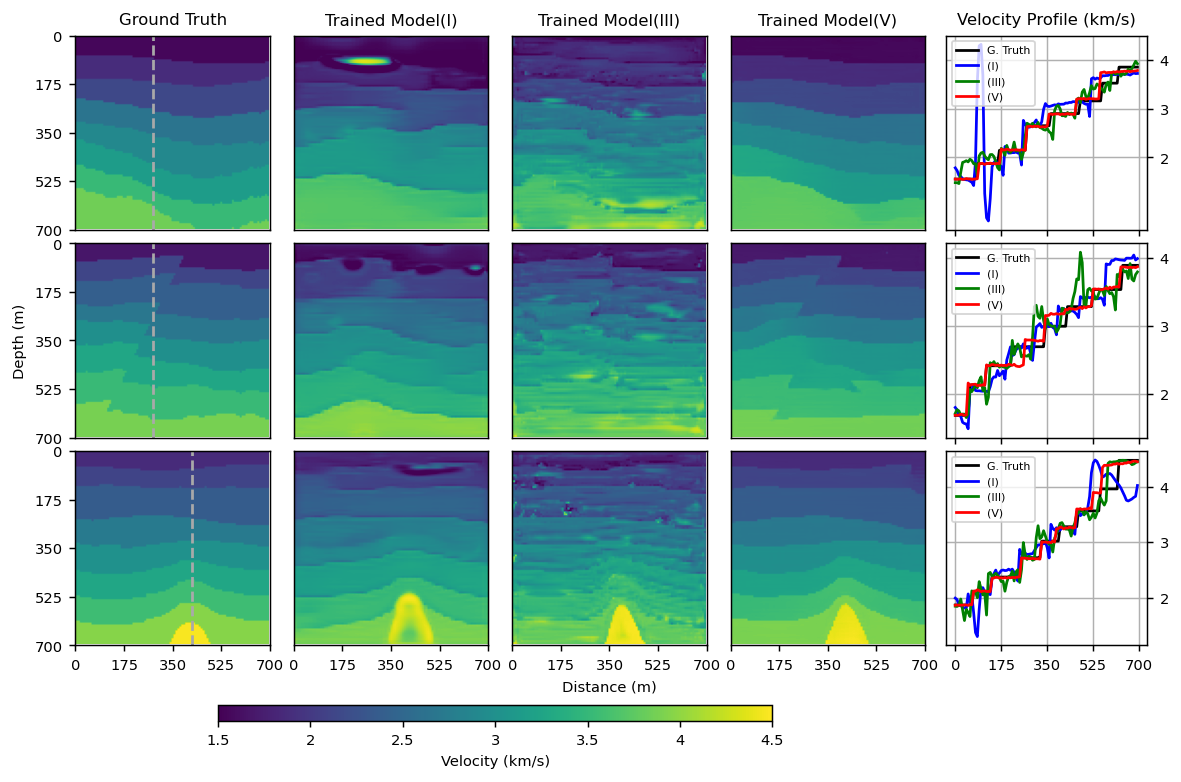}
\caption{Prediction outcomes of our models for the noisy dataset. The first column corresponds to the ground truth (target) velocity model, while the rightmost column displays a vertical velocity profile along the gray line depicted in the target model for four velocity models. The velocity models are the same as Fig. \ref{fig-Noisles-SVInvNet}. Best viewed in the electronic version with zoom.}
\label{fig-Nois-SVInvNet}
\end{figure}
%------

As the depth increases, predicting the shape of the layer interfaces and velocity values becomes increasingly challenging. From a geophysical standpoint, as the seismic wave propagates through various subsurface layers, a portion of its amplitude and thus its energy is lost due to reflection back to the Earth's surface. The decrease of energy gives rise to reduction in the amplitude of seismic data reflected from deeper interfaces, particularly in cases characterized by a lower reflection coefficient. Thereby, as the number of layers increases, seismic data originating from deep interfaces exhibit decreased amplitudes and insufficient  information. Hence, the amplitude of the data recorded by receivers diminishes over time, posing a challenge for CNN kernels in the extraction of features from seismic data. In Fig. \ref{fig-Noisles-SVInvNet}, in the first-row, Model-\RomanNumeralCaps{1} encountered challenges in constructing the 7th layer of the velocity model and resorted to distributing this layer between the 6th and 8th layers. Conversely, Model-\RomanNumeralCaps{3} modified the shapes of the 6th and 8th layers, and constructed a portion of the 7th layer, as indicated by the yellow arrow. Model-\RomanNumeralCaps{5}, in contrast, could successfully constructed the 7th layer.

In fault models, the presence of abrupt angles along the fault lines perturbs the trajectory of the reflective hyperbola. Concurrently, diffractions arising at proximal surface interfaces result in the attenuation of reflections and diffractions from deeper strata, a phenomenon that becomes particularly evident when the displacement of the fault, or fault throw, is relatively short.

This complexity is amplified particularly in the case of faults located in close proximity to the edges. As the quantity of training dataset samples increased, the deep model demonstrated an enhanced ability to successfully construct the shapes of fault lines and faulty interfaces. The white arrows delineate the actual position of the fault line in the velocity models, facilitating ease of comparison.

The salt dome has distinctive properties in both velocity value and shape, different from the surrounding layers. Seismic waves encountering the apex of a salt dome undergo scattering, thereby creating a noticeable scattering pattern. However, due to its placement in the lower layer, the amplitude of these scatterings is considerably smaller when compared to the reflection data. As observed in the third row of Fig. \ref{fig-Noisles-SVInvNet}, the deep model trained with a lower number of training data identified the scatterings originating from the layers above the salt dome as belonging to a large salt body; with an increased number of training samples, the other two models exhibits an improved capability to discern the structural characteristics of the salt dome and differentiate the scattering caused by it from the surrounding layers.

%----- figure 10
\begin{figure}[!t]
\centering
\includegraphics[width=0.48\textwidth]{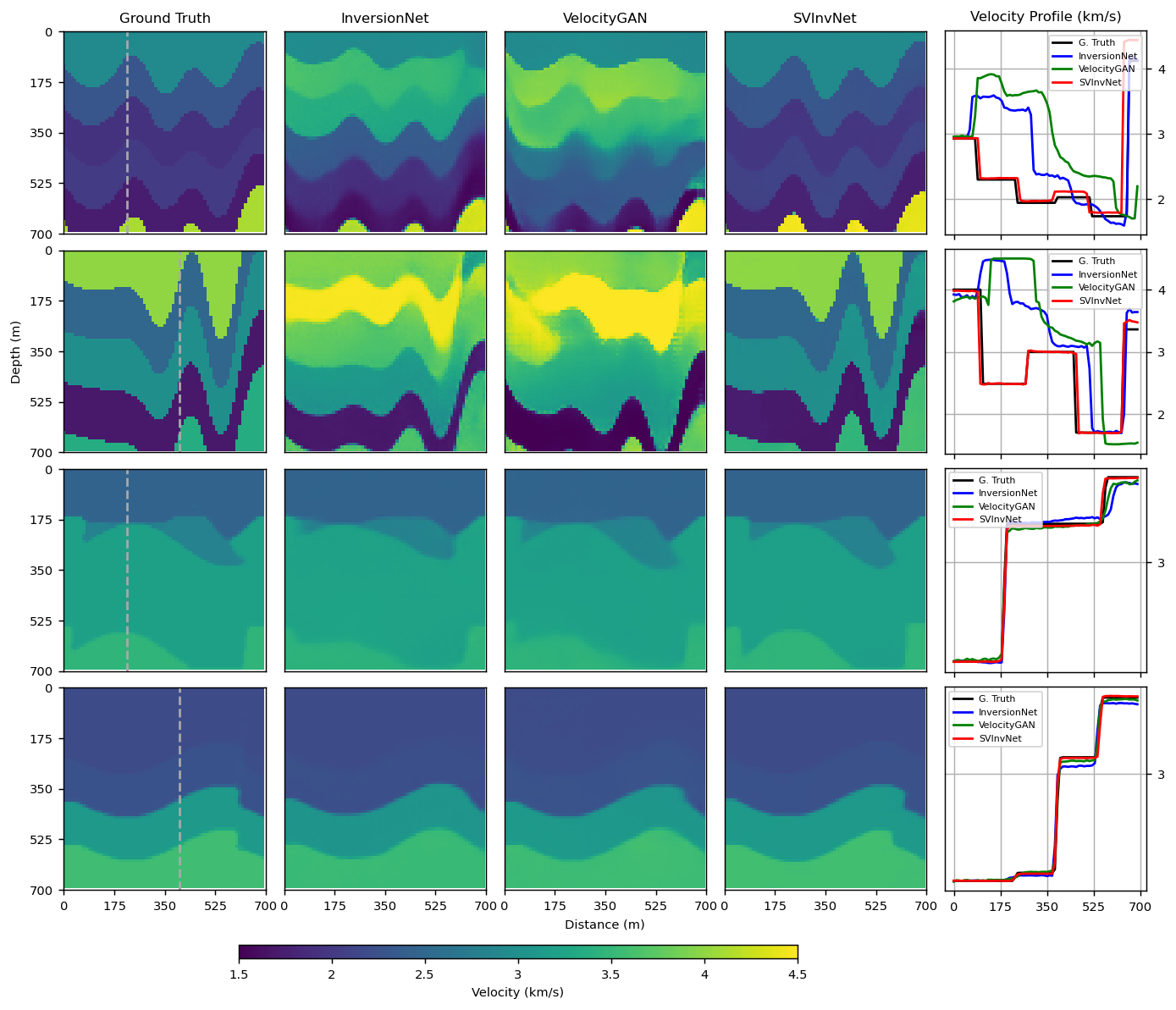}
\caption{Sample ground truth velocity models and their prediction results of InversionNet, VelocityGAN, and SVInvNet for a) CurveVel-B and b) CurveFault-A benchmarks, with the vertical velocity profiles on the right-most column. The velocity profiles belong to the gray lines depicted in the corresponding ground truth models. Best viewed in the electronic version with zoom.}
\label{fig-CV-CF}
\end{figure}
%------

%---- Table
\begin{table} [!b]
\begin{center}
\caption{The Performance of Four DNN Models on the Test Set. Models are trained utilizing CurveVel-B (CV-B) and CurveFault-A (CF-A) datasets.}
\label{table:OpenFWI-SVInvNet}
\begin{tabular}{c c c c c c c}
\hline
& \multicolumn{3}{|c}{CV-B} & \multicolumn{3}{|c}{CF-A} \\ 
\hline \hline
Metric & \multicolumn{1}{|c}{MAE}& RMSE & \multicolumn{1}{c|}{SSIM}& MAE & RMSE & SSIM\\ 
\hline
\multicolumn{1}{c|}{SVInvNet} & \textbf{0.0250} & \textbf{0.0583} &\multicolumn{1}{c|}{\textbf{0.9997}} & \textbf{0.0042} & \textbf{0.0128} & \textbf{0.9999} \\

\multicolumn{1}{c|}{InversionNet} & 0.1448 & 0.3111 &\multicolumn{1}{c|}{0.6630} & 0.0303 & 0.0766 & 0.9448 \\

\multicolumn{1}{c|}{VelocityGAN} & 0.1268 & 0.2618 &\multicolumn{1}{c|}{0.7111} & 0.0216 & 0.0505 & 0.9687 \\

\multicolumn{1}{c|}{UPFWI} & 0.1777 & 0.3179 &\multicolumn{1}{c|}{0.6614} & 0.0500 & 0.0966 & 0.9495 \\

\hline
\end{tabular}
\end{center}
\end{table}
%-----
The introduction of noise to the data exacerbates this situation, causing distortion in the amplitude of the signal emanating from the interfaces and deforming the trajectory of the scattering or reflection hyperbola. This further complicates the process of extracting features from the seismic gathers performed by CNN kernels. The adverse impact of noise in the velocity mapping process is evident based on the performances presented in Tables \ref{table:ourmodels}. In the Table \ref{table:ourmodels}, it is evident that SVInvNet demonstrates superior performance, even in the presence of noise, when compared to the InversionNet.

Fig. \ref{fig-Nois-SVInvNet}, depicts the predicted velocity values of SVInvNet trained with noisy TD-\RomanNumeralCaps{1}, TD-\RomanNumeralCaps{3}, and TD-\RomanNumeralCaps{5}.  The performances of Model-\RomanNumeralCaps{1} and Model-\RomanNumeralCaps{3} show significantly inferior results compared to the corresponding noiseless cases. The prediction results presented in Table \ref{table:ourmodels} substantiate this. To successfully construct the velocity model by capturing the signal within noisy data and discerning its pattern, the deep model requires a greater number of samples for supervision. In other words, an increased sample size is imperative for the model to effectively filter out noise and accurately learn the desired signal. 

\subsubsection{Analysis Using the OpenFWI Dataset}
The performance evaluation of the SVInvNet deep model is conducted using the open source dataset OpenFWI, as published by Deng et al. \cite{NEURIPS2022_27d3ef26}. They trained three distinct deep models utilizing various subgroups of OpenFWI and report the results. These deep models include InversionNet \cite{Wu}, VelocityGAN \cite{velGAN}, and UPFWI \cite{jin2022unsupervised}. We employed the subset benchmarks of OpenFWI, namely the CurveVel-B and CurveFault-A datasets, for both the training and testing phases of the SVInvNet.

Compared to our dataset, the OpenFWI dataset includes 70 receivers and uses velocity models sized at 70$\times$70. To match this, we add an extra layer to the SVInvNet first four layers and change the final layers' size from 100$\times$100 to 70$\times$70.  It was noted to emphasize that both training and testing were conducted on the same benchmark, as previously discussed. Therefore, it was removed. We compare our results with the best values from their study, as shown in Table \ref{table:OpenFWI-SVInvNet}. The data in Table \ref{table:OpenFWI-SVInvNet} clearly shows that our model performs better than the other deep models. We also utilized the corresponding pretrained models of InversionNet and VelocityGAN to provide visual comparisons with SVInvNet, using samples randomly selected from both the CurveVel-B and CurveFault-A test datasets, as presented in Fig. \ref{fig-CV-CF}. The samples in Fig. \ref{fig-CV-CF} align with the quantitative values presented in Table \ref{table:OpenFWI-SVInvNet}. SVInvNet demonstrated superior performance in predicting both the velocity values and the interface shapes when compared to the other models.

%----- figure 11
\begin{figure*}[!t]
\renewcommand\thefigure{11}
\centering
\includegraphics[width=0.6\textwidth]{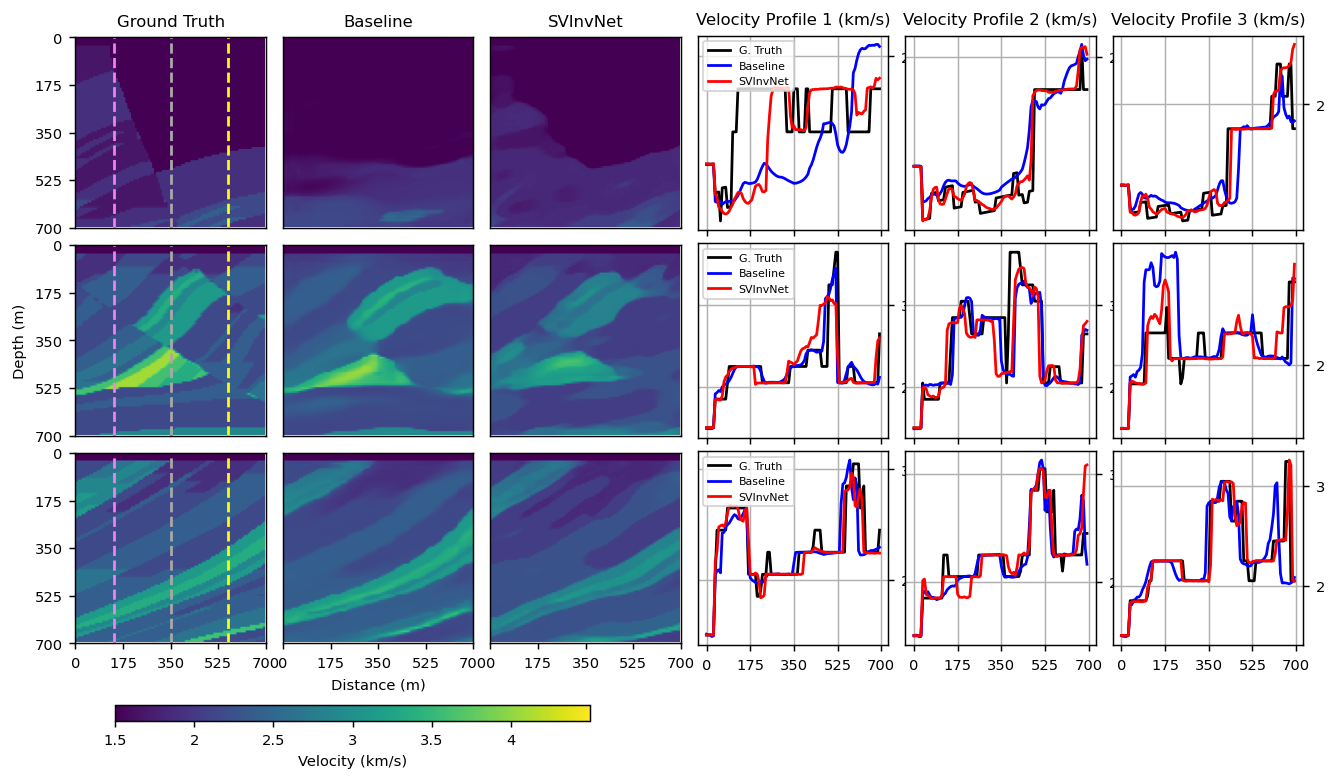}
\caption{Sample ground truth velocity models and the corresponding prediction results from the baseline model and SVInvNet for the Marmousi benchmarks, with vertical velocity profiles shown in the right-most column. The velocity profiles correspond to the violet, gray, and yellow lines as shown in their respective ground truth models. }
\label{Marmousi}
\end{figure*}
%------

\subsubsection{Analysis Using the Marmousi-based models}
To assess the performance of the SVInvNet, the models obtained from the Marmousi model was utilized, as it provides a more accurate representation of the true velocity distribution within the subsurface. To accomplish this, we generated 407 sub-model sets by systematically scanning the Marmousi velocity model with a 200×200 template at predetermined intervals. We ensured that the velocity contrast in the prepared models was at least 200 m/s in the vertical direction. The models were then downsampled to a 100×100 size. We randomly selected 23 models from the dataset for testing, while the remaining 384 models were combined with our own TD-\RomanNumeralCaps{5} to train the baseline model and SVInvNet. The trained models was then tested on the 23 test datasets.  Qualitative evaluations of the results for randomly selected test samples are presented in Fig. \ref{Marmousi}. As demonstrated in the Fig. \ref{Marmousi}, SVInvNet outperforms the baseline; however, the training dataset is inadequate for accurately estimating high-slope faulted transitions and nested sloped layers with low contrast.

\section{Conclusion}
\label{sec:Con}
In this study, we introduce a new end-to-end deep model, SVInvNet, for seismic velocity inversion. In order to enhance the capacity of SVInvNet in learning and estimating more complex velocity models, we abstained from merely increasing the channel number of CNN layers (i.e. number of feature planes), which cause an excessive increase in the number of parameters. Instead, we devised a network architecture characterized by the multiple connections among its layers with only 4M parameters. This design facilitated the efficient flow of information between the input and the corresponding ground truth. Based on the outcomes derived from both noisy and noiseless datasets, SVInvNet demonstrates proficiency in accurately estimating velocity models characterized by multilayer structures, fault, and salt dome. Furthermore, it is observed that increasing the quantity of training dataset correlates with a reduction in loss function in the testing phase. The utilization of a test dataset size at least twice that of the training datasets reveals the comprehensiveness of the results produced by the trained SVInvNet.

The results obtained from the dataset containing both random and coherent noise emphasize the capability of SVInvNet to effectively filter such noise types and map seismic data to velocity models. However, attaining an equivalent loss function value as observed in the noise-free dataset necessitates a larger training dataset in the presence of noise.

%\begin{IEEEbiographynophoto}{Muctafa Necefi}
%Biography text here without a photo.
%\end{IEEEbiographynophoto}

%\begin{IEEEbiography}%[{\includegraphics[width=1in,height=1.25in,clip,keepaspectratio]{fig1.png}}]{IEEE Publications Technology Team}
%In this paragraph you can place your educational, professional background and research and other interests.\end{IEEEbiography}

%\input{bib.tex}

%\section{Acknowledgments}
%\noindent We acknowledge the assistance of OpenAI's ChatGPT-4 for editorial help in refining the language and improving the clarity of this manuscript. It is important to clarify that no part of the text was solely generated by the AI; rather, ChatGPT served as a tool for enhancing the clarity and readability of the content originally created by the authors.

\bibliographystyle{IEEEtran}
\bibliography{references}

\begin{IEEEbiography}[{\includegraphics[width=1in,height=1.25in,clip,keepaspectratio]{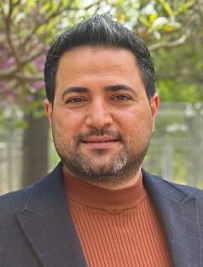}}]{Mojtaba Najafi Khatounabad} received the bachelor’s degree in Nuclear Physics and the master’s degree in Geophysics from Urmia University, Urmia, Iran, in 2010, and 2014 respectively. In 2017, he was awarded a full Ph.D. scholarship by Türkiye Scholarships. He is currently pursuing the Ph.D. degree in Engineering Geophysics with Institute of Science and Technology, Ankara University, Ankara, Türkiye, under the supervision of Prof. Dr. Selma Kadioglu and Assoc. Prof. Dr. Hacer Yalim Keles. 
Najafi Khatounabad is currently involved in a project focused on developing the program for seismic data processing. 

His research interest includes seismic modeling, full-waveform inversion, and deep-learning-based geophysical data processing and inversion.
\end{IEEEbiography}
\vspace{-10mm}
\begin{IEEEbiography}[{\includegraphics[width=1in,height=1.25in,clip,keepaspectratio]{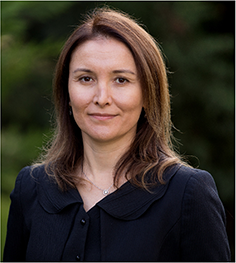}}]{Dr. Hacer Yalim Keles} completed her B.Sc., M.Sc., and Ph.D. in Computer Engineering at Middle East Technical University, Turkey, in 2002, 2005, and 2010, respectively. Her distinguished Ph.D. thesis was honored with the Thesis of the Year award by the Prof. Dr. Mustafa Parlar Education and Research Foundation in 2010. Between 2000 and 2007, she contributed as a researcher at The Scientific and Technological Research Council of Turkey (TUBITAK), focusing on pattern recognition using multimedia data, including audio and video.

In 2010, she founded her own R\&D company with a grant from the Ministry of Industry and Trade of Turkey. Her project SOYA, funded by TUBITAK in 2011, was later recognized as one of the best venture projects, leading to an opportunity in Silicon Valley for potential investments. Dr. Keles was an Assistant Professor at the Department of Computer Engineering, Ankara University, from 2013 to 2021, and is currently an Associate Professor at the Department of Computer Engineering, Hacettepe University.

Her research primarily spans computer vision and machine learning, with a focus on learning algorithms for limited data and deep generative models. She has contributed to sign and gesture recognition, generative adversarial networks, image inpainting, and image segmentation domains. Moreover, she collaborates on diverse projects involving aerial and medical images, speech signals, textual, geophysical, and hyperspectral data analysis with her graduate students.

\end{IEEEbiography}
\vspace{-10mm}
\begin{IEEEbiography}[{\includegraphics[width=1in,height=1.25in,clip,keepaspectratio]{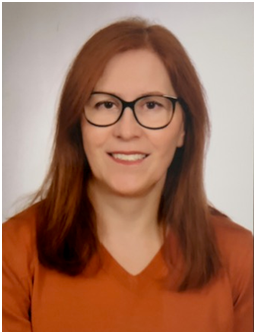}}]{Prof. Selma Kadioglu } completed her B.Sc. in Geophysical Engineering at Black Sea Technical University, Trabzon-Turkey, M.Sc. and Ph.D. in Geophysical Engineering at Ankara University, supervised by Prof. Turan KAYIRAN, Ankara-Turkey, in 1990, 1993, and 2000, respectively. During her undergraduate studies, she was awarded as a high honours student and graduated as the second runner-up of the university. She focused on seismic wavefield computation methods during M.Sc. and Ph.D. and developed a series approximation of the Kirchhoff integral method for the rough subsurface models. While doing her Ph.D., she received the Scientific and Technological Research Council of Turkey (TUBITAK) Münir Birsel Foundation doctoral research fellowship between 1996-1998. At the same time, she received the Council of Higher Education of Turkey (YOK) funded overseas doctoral research fellowship between 1997-1999 and she studied with Prof. Dan Kosloff about seismic modeling and tomography of depth migrated gathers for transversely isotropic medium, in Department of Geophysics and Planetary Sciences of Tel Aviv University, Tel-Aviv-Israel. Then she studied with Prof. Jeffrey J. Daniels in School of Earth Sciences of the Ohio-State University, Ohio-USA, about ground penetrating radar (GPR) during 10 months in 2001 and developed a new 3D visualisation of integrated GPR data and EM-61 data to determine buried objects and their characteristics with him. Prof. Kadioglu is currently a Professor at the Department of Geophysical Engineering, Ankara University. Her research interests include transparent/semi-transparent half bird's eye view visualisation of 3D GPR data volume, seismic and GPR forward and inverse wavefield scattering modeling, tomography and data processing, archaeogeophysics, determining geological structures such as fractures, faults, cavities, buried infrastructures, stability problems of cultural constructions and mining geophysics.

\end{IEEEbiography}

\end{document}